\documentclass[12pt,english]{article}
\usepackage[LGR,T1]{fontenc}
\usepackage[latin9]{inputenc}
\usepackage[a4paper]{geometry}
\usepackage{fancyhdr}
\usepackage{color}
\usepackage{babel}
\usepackage{float}
\usepackage{url}
\usepackage{amsmath}
\usepackage{graphicx}
\usepackage{wasysym}
\usepackage[unicode=true,pdfusetitle,
 bookmarks=true,bookmarksnumbered=false,bookmarksopen=false,
 breaklinks=false,pdfborder={0 0 1},backref=false,colorlinks=false]
 {hyperref}
\hypersetup{
 colorlinks=true,citecolor=blue}

\makeatletter

\DeclareRobustCommand{\greektext}{%
  \fontencoding{LGR}\selectfont\def\encodingdefault{LGR}}
\DeclareRobustCommand{\textgreek}[1]{\leavevmode{\greektext #1}}
\ProvideTextCommand{\~}{LGR}[1]{\char126#1}

\providecommand{\tabularnewline}{\\}
\floatstyle{ruled}
\newfloat{algorithm}{tbp}{loa}
\providecommand{\algorithmname}{Algorithm}
\floatname{algorithm}{\protect\algorithmname}

\newenvironment{lyxcode}
	{\par\begin{list}{}{
		\setlength{\rightmargin}{\leftmargin}
		\setlength{\listparindent}{0pt}
		\raggedright
		\setlength{\itemsep}{0pt}
		\setlength{\parsep}{0pt}
		\normalfont\ttfamily}%
	 \item[]}
	{\end{list}}

\usepackage{lastpage}
\usepackage{breakcites}
\usepackage{indentfirst}

\makeatother

\begin{document}
\title{Exponential Kernels with Latency in Hawkes Processes: Applications
in Finance}
\date{Jan-2021}
\author{Marcos Costa Santos Carreira\thanks{École Polytechnique, CMAP - PhD under the Quantitative Regulation
chair}}
\maketitle
\begin{abstract}
The Tick library \cite{tick} allows researchers in market microstructure
to simulate and learn Hawkes process in high-frequency data, with
optimized parametric and non-parametric learners. But one challenge
is to take into account the correct causality of order book events
considering latency: the only way one order book event can influence
another is if the time difference between them (by the central order
book timestamps) is greater than the minimum amount of time for an
event to be (i) published in the order book, (ii) reach the trader
responsible for the second event, (iii) influence the decision (processing
time at the trader) and (iv) the 2nd event reach the order book and
be processed. For this we can use exponential kernels shifted to the
right by the latency amount. We derive the expression for the log-likelihood
to be minimized for the 1-D and the multidimensional cases, and test
this method with simulated data and real data. On real data we find
that, although not all decays are the same, the latency itself will
determine most of the decays. We also show how the decays are related
to the latency. Code is available on GitHub at \url{https://github.com/MarcosCarreira/Hawkes-With-Latency}.


\pagebreak{}
\end{abstract}

\section{Introduction}

It is well known that markets function in response both to external
(or exogenous) information - e.g. news - and internal (endogenous)
information, like the behavior of market participants and the patterns
of price movements. The systematization of markets (synthesized by
the Central Limit Order Book) led to two important developments on
these responses: order book information is now available in an organized
and timely form (which enables the study of intraday patterns in historical
data) and algorithms can be deployed to process all kinds of information
fast enough to trade (even before the information is received or processed
by all the market participants).

That combination increases the importance of understanding how endogenous
trading interacts with itself: activity in the order book leads to
activity from fast traders which leads to more activity and so on.
While the arrival of orders in an order book due to exogenous information
can be described as a Poisson process, the interaction of the agents
correspond to a self-excitatory process, which can be described by
Hawkes processes. The importance of Hawkes processes in finance has
been described quite well in \cite{Bacry et al 2015}, and recent
work by \cite{Euch et al 2016} has established the connection between
Hawkes processes, stylized market microstructure facts and rough volatility.
More recently, \cite{Bacry et al 2019} improve the Queue-Reactive
model by \cite{Huang et al 2015} by adding Hawkes components to the
arrival rates.

We start focusing on \cite{Bacry et al 2015}, which apply Hawkes
processes learners as implemented in the Tick library \cite{tick}
to futures. There are two limitations in these applications: for parametric
learners the decay(s) must be given and be the same for all kernels,
and influences are assumed to be instantaneous. In this short paper
we show how to code a parametric learner for an exponential kernel
(or kernels in the multivariate case) in which we learn the decay
and consider a latency (given). We won't revisit the mathematical
details for the known results; for this, we refer the reader to \cite{Toke 2011}
and \cite{Abergel et al 2016}. One of the goals is to show a working
Python script optimized at the most by \cite{Numba}; this can be
optimized for improved performance. The other goal is to understand
what the learned parameters of the kernels mean given the trading
dynamics of the BUND futures.

\section{One-dimensional Hawkes processes}

\subsection{Definition}

As explained in \cite{Toke 2011}, a Hawkes process with an exponential
kernel has its intensity described by:

\begin{equation}
\lambda\left(t\right)=\lambda_{0}\left(t\right)+\sum_{t_{i}<t}\sum_{j=1}^{P}\left[\alpha_{j}\cdot\exp\left(-\beta_{j}\cdot\left(t-t_{i}\right)\right)\right]
\end{equation}

For $P=1$ and $\lambda_{0}\left(t\right)=\lambda_{0}$ :

\begin{equation}
\lambda\left(t\right)=\lambda_{0}+\sum_{t_{i}<t}\left[\alpha\cdot\exp\left(-\beta\cdot\left(t-t_{i}\right)\right)\right]
\end{equation}
With stationarity condition:

\begin{equation}
\frac{\alpha}{\beta}<1
\end{equation}
And unconditional expected value of intensity:

\begin{equation}
E\left[\lambda\left(t\right)\right]=\frac{\lambda_{0}}{1-\frac{\alpha}{\beta}}
\end{equation}

\subsection{Maximum-likelihood estimation}

Following \cite{Toke 2011}:

\begin{equation}
LL=\varint_{0}^{T}\left(1-\lambda\left(s\right)\right)ds+\varint_{0}^{T}\left(\ln\left(\lambda\left(s\right)\right)\right)dN\left(s\right)
\end{equation}

Which is computed as:

\begin{equation}
LL=t_{N}-\varint_{0}^{T}\lambda\left(s\right)ds+\sum_{i=1}^{N}\ln\left[\lambda_{0}\left(t_{i}\right)+\sum_{j=1}^{P}\sum_{k=1}^{i-1}\alpha_{j}\cdot\exp\left(-\beta_{j}\cdot\left(t_{i}-t_{k}\right)\right)\right]
\end{equation}

Defining the function:

\begin{equation}
R_{j}\left(i\right)=\sum_{k=1}^{i-1}\exp\left(-\beta_{j}\cdot\left(t_{i}-t_{k}\right)\right)
\end{equation}

The recursion observed by \cite{Ogata et al 1981} is:

\begin{equation}
R_{j}\left(i\right)=\exp\left(-\beta_{j}\cdot\left(t_{i}-t_{i-1}\right)\right)\cdot\left(1+R_{j}\left(i-1\right)\right)
\end{equation}

And therefore the log-likelihood is:

\begin{align}
LL & =t_{N}-\varint_{0}^{T}\lambda_{0}\left(s\right)ds-\sum_{i=1}^{N}\sum_{j=1}^{P}\left[\frac{\alpha_{j}}{\beta_{j}}\left(1-\exp\left(-\beta_{j}\cdot\left(t_{N}-t_{i}\right)\right)\right)\right]\nonumber \\
 & +\sum_{i=1}^{N}\ln\left[\lambda_{0}\left(t_{i}\right)+\sum_{j=1}^{P}\alpha_{j}\cdot R_{j}\left(i\right)\right]
\end{align}

For $P=1$ and $\lambda_{0}\left(t\right)=\lambda_{0}$ :

\begin{equation}
R\left(i\right)=\exp\left(-\beta\cdot\left(t_{i}-t_{i-1}\right)\right)\cdot\left(1+R\left(i-1\right)\right)
\end{equation}

And:

\begin{align}
LL & =t_{n}-\lambda_{0}\cdot T-\sum_{i=1}^{N}\left[\frac{\alpha}{\beta}\left(1-\exp\left(-\beta\cdot\left(t_{N}-t_{i}\right)\right)\right)\right]\nonumber \\
 & +\sum_{i=1}^{N}\ln\left[\lambda_{0}+\alpha\cdot R\left(i\right)\right]
\end{align}

\subsection{Simulation and learning}

Using tick we simulate 100 paths for different end times (100, 1000,
10000) for both the builtin \textit{ExpKernel} and for a user-defined
Time Function that mirrors an \textit{ExpKernel}:

\begin{algorithm}[H]
\begin{lyxcode}
Initialize~$\left\{ \lambda_{0},\alpha,\beta\right\} $

Define~the~kernel~function~$f\left(\alpha,\beta,t\right)$

Define~the~support~$s$

Build~the~\textit{TimeFunction}~with:
\begin{lyxcode}
n~+~1~steps~for~the~time~interval~$\left\{ 0,s\right\} $

n~+~1~evaluations~of~f~for~these~points

an~interpolation~mode~(e.g.~\textit{TimeFunction.InterConstRight})
\end{lyxcode}
Build~a~Kernel~with~\textit{HawkesKernelTimeFunc}

Define~the~builtin~\textit{HawkesKernelExp}~with~parameters~$\left\{ \frac{\alpha}{\beta},\beta\right\} $

Define~the~number~of~simulations~$N_{S}$

Define~the~end~time~$T$

Build~the~\textit{SimuHawkes}~object~with~$\lambda_{0}$,~a~kernel~and~$T$

Build~the~\textit{SimuHawkesMulti}~object~with~\textit{SimuHawkes}~and~$N_{S}$

Simulate~\textit{SimuHawkesMulti}
\end{lyxcode}
\caption{Definitions of Exponential Kernels (builtin and custom) and simulation}
\end{algorithm}

We use \textit{minimize} from \cite{scipy} with method \textcolor{red}{'SLSQP'};
as pointed out in \cite{Kisel 2017}, it can handle both bounds (ensuring
positivity of parameters) and constraints (ensuring the stationarity
condition). The differences should be calculated at the start of the
optimization process and passed to the likelihood function instead
of being calculated inside the likelihood function.

\begin{algorithm}[H]
\begin{lyxcode}
def~$ll\left(\theta,ts,\Delta ts,\delta ts\right)$:
\begin{lyxcode}
$\left\{ \lambda_{0},\alpha,\beta\right\} =\theta$~parameters~are~defined~as~one~argument~for~the~optimization

$ts$~is~one~of~the~\textit{SimuHawkesMulti.timestamps}

$\Delta ts$~is~$ts_{i}-ts_{i-1}$

$\delta ts$~is~$ts_{N}-ts_{i-1}$

Define~$R$~as~an~array~of~zeros~with~the~same~size~as~$ts$

Keep~its~first~element~$R_{1}=0$

Calculate~$R_{i}=\exp\left(-\beta\cdot\Delta ts_{i-1}\right)\cdot\left(1+R_{i-1}\right)$~recursively~for~$i>1$

Return:

$-\left[ts_{N}\cdot\left(1-\lambda_{0}\right)-\left(\frac{\alpha}{\beta}\right)\cdot\sum_{\delta ts}\left(1-\exp\left(-\beta\cdot\delta ts\right)\right)+\sum_{R_{i}}\left(\log\left(\lambda_{0}+\alpha\cdot R_{i}\right)\right)\right]$
\end{lyxcode}
-{}-{}-{}-{}-{}-{}-{}-{}-{}-{}-{}-{}-{}-{}-{}-{}-{}-{}-{}-{}-{}-{}-{}-{}-{}-{}-{}-{}-{}-{}-{}-{}-{}-{}-{}-{}-{}-{}-{}-{}-{}-{}-{}-{}-{}-{}-{}-{}-{}-{}-{}-{}-{}-{}-{}-{}-{}-{}-{}-{}-{}-{}-{}-{}-{}-{}-{}-{}-{}-{}-{}-{}-{}-{}-{}-

Minimize~$ll\left(\theta,ts,\Delta ts,\delta ts\right)$~with:
\begin{lyxcode}
bounds~on~$\lambda_{0},\alpha,\beta$~(all~must~be~positive)~and

constraint~$\alpha<\beta$
\end{lyxcode}
\end{lyxcode}
\caption{Negative Log-Likelihood and its minimization with scipy's minimize}
\end{algorithm}

\subsection{Kernels with latency}

For $P=1$ and $\lambda_{0}\left(t\right)=\lambda_{0}$ , with latency
\textgreek{t}:

\begin{equation}
\lambda\left(t\right)=\lambda_{0}+\sum_{t_{i}<t-\tau}\left[\alpha\cdot\exp\left(-\beta\cdot\left(t-\tau-t_{i}\right)\right)\right]
\end{equation}

We can use the same (exponential) function and shift it to the right
by the latency:

\begin{algorithm}[H]
\begin{lyxcode}
Initialize~$\left\{ \lambda_{0},\alpha,\beta\right\} $

Define~the~kernel~function~$f\left(\alpha,\beta,t\right)$

Define~the~support~$s$

Build~the~\textit{TimeFunction}~with:
\begin{lyxcode}
n~+~1~steps~for~the~time~interval~$\left\{ 0,s\right\} $

n~+~1~evaluations~of~f~for~these~points

an~interpolation~mode~(e.g.~\textit{TimeFunction.InterConstRight})

If~latency~$\tau>0$:
\begin{lyxcode}
shift~$\left\{ 0,s\right\} $~and~$f\left(\left\{ 0,s\right\} \right)$~to~the~right~by~$\tau$

add~zeros~to~the~interval~$\left\{ 0,\tau\right\} $
\end{lyxcode}
\end{lyxcode}
Build~a~Kernel~with~\textit{HawkesKernelTimeFunc}

Define~the~number~of~simulations~$N_{S}$

Define~the~end~time~$T$

Build~the~\textit{SimuHawkes}~object~with~$\lambda_{0}$,~the~kernel~and~$T$

Build~the~\textit{SimuHawkesMulti}~object~with~\textit{SimuHawkes}~and~$N_{S}$

Simulate~\textit{SimuHawkesMulti}
\end{lyxcode}
\caption{Exponential Kernel with latency}
\end{algorithm}

The new likelihoods are defined (for $P=1$ and $\lambda_{0}\left(t\right)=\lambda_{0}$)
as:

\begin{align}
R_{\tau}\left(i\right) & =R_{\tau}\left(i-1\right)\cdot\exp\left(-\beta\cdot\left(t_{i}-t_{i-1}\right)\right)\\
 & +\begin{cases}
\sum_{t_{k}}\left[\exp\left(-\beta\cdot\left(t_{i}-\tau-t_{k}\right)\right)\right] & t_{i-1}-\tau\leq t_{k}<t_{i}-\tau\\
0 & otherwise
\end{cases}
\end{align}

With $R_{\tau}\left(1\right)=0$, and:

\begin{align}
LL_{\delta} & =t_{N}-\lambda_{0}\cdot t_{N}-\sum_{t_{i}<t-\tau}\left[\frac{\alpha}{\beta}\left(1-\exp\left(-\beta\cdot\left(t_{N}-\tau-t_{i}\right)\right)\right)\right]\nonumber \\
 & +\sum_{i=1}^{n}\ln\left[\lambda_{0}+\alpha\cdot R_{\tau}\left(i\right)\right]
\end{align}

Even with the latency $\tau=0$, for each $t_{i}$ we pass $t_{i-1}$.
But this enough to slow down the optimization, as observed while testing
that the algorithm returns the same results as the previous one for
latency zero and seen in Table \ref{tab:Learning-results-and}.

\begin{algorithm}[H]
\begin{lyxcode}
def~$lllat\left(\theta,t,\Delta t,\delta t,\tau t\right)$:
\begin{lyxcode}
$\left\{ \lambda_{0},\alpha,\beta\right\} =\theta$~parameters~are~defined~as~one~argument~for~the~optimization

$t$~is~one~of~the~\textit{SimuHawkesMulti.timestamps}

$\Delta t$~is~the~series~$t_{i}-t_{i-1}$~(excluding~$t_{1}$)

$\delta t$~is~the~series~$t_{N}-\tau-t_{i-1}$

$\tau t$~is~the~series~of~arrays~$t_{i}-\tau-t_{k}$~for~all~$t_{k}$~such~that~\textrm{$t_{i-1}-\tau\leq t_{k}<t_{i}-\tau$}

Define~$R$~as~an~array~of~zeros~with~the~same~size~as~$ts$

Keep~its~first~element~$R_{1}=0$

Calculate~$R_{i}=\exp\left(-\beta\cdot\Delta t_{i-1}\right)\cdot R_{i-1}+\sum_{\tau t_{i}}\left(\exp\left(-\beta\cdot\tau t_{i}\right)\right)$~recursively~for~$i>1$

Return:

$-\left[t_{N}\cdot\left(1-\lambda_{0}\right)-\left(\frac{\alpha}{\beta}\right)\cdot\sum_{\delta t}\left(1-\exp\left(-\beta\cdot\delta t\right)\right)+\sum_{R_{i}}\left(\log\left(\lambda_{0}+\alpha\cdot R_{i}\right)\right)\right]$
\end{lyxcode}
-{}-{}-{}-{}-{}-{}-{}-{}-{}-{}-{}-{}-{}-{}-{}-{}-{}-{}-{}-{}-{}-{}-{}-{}-{}-{}-{}-{}-{}-{}-{}-{}-{}-{}-{}-{}-{}-{}-{}-{}-{}-{}-{}-{}-{}-{}-{}-{}-{}-{}-{}-{}-{}-{}-{}-{}-{}-{}-{}-{}-{}-{}-{}-{}-{}-{}-{}-{}-{}-{}-{}-{}-{}-{}-{}-

Minimize~$lllat\left(\theta,t,\Delta t,\delta t,\tau t\right)$~with:
\begin{lyxcode}
bounds~on~$\lambda_{0},\alpha,\beta$~(all~must~be~positive)~and

constraint~$\alpha<\beta$
\end{lyxcode}
\end{lyxcode}
\caption{Log-likelihood with latency}

\end{algorithm}

We obtain the following results for 100 paths (parameters $\lambda_{0}=1.20$
, $\alpha=0.60$ and $\beta=0.80$) with the total running times in
seconds (on a MacBook Pro 16-inch 2019, 2.4 GHz 8-core Intel Core
i9, 64 GB 2667 MHz DDR4 RAM):

\begin{table}[H]
\begin{centering}
\begin{tabular}{|r|c|c|r@{\extracolsep{0pt}.}l|r@{\extracolsep{0pt}.}l|r@{\extracolsep{0pt}.}l|r@{\extracolsep{0pt}.}l|}
\hline 
End Time & Kernel & Algorithm & \multicolumn{2}{c|}{runtime (s)} & \multicolumn{2}{c|}{$\lambda_{0}$} & \multicolumn{2}{c|}{$\alpha$} & \multicolumn{2}{c|}{$\beta$}\tabularnewline
\hline 
\hline 
100 & Builtin & \textit{ll} & 0&535 & 1&40 & 0&59 & 0&87\tabularnewline
\hline 
100 & Builtin & \textit{lllat} & 1&49 & 1&40 & 0&59 & 0&87\tabularnewline
\hline 
100 & Custom & \textit{ll} & 0&521 & 1&41 & 0&63 & 0&88\tabularnewline
\hline 
100 & Custom & \textit{lllat} & 1&6 & 1&41 & 0&63 & 0&88\tabularnewline
\hline 
1000 & Builtin & \textit{ll} & 0&992 & 1&22 & 0&60 & 0&80\tabularnewline
\hline 
1000 & Builtin & \textit{lllat} & 11&3 & 1&22 & 0&60 & 0&80\tabularnewline
\hline 
1000 & Custom & \textit{ll} & 1&13 & 1&23 & 0&63 & 0&81\tabularnewline
\hline 
1000 & Custom & \textit{lllat} & 12&8 & 1&23 & 0&63 & 0&81\tabularnewline
\hline 
10000 & Builtin & \textit{ll} & 6&02 & 1&20 & 0&60 & 0&80\tabularnewline
\hline 
10000 & Builtin & \textit{lllat} & 143& & 1&20 & 0&60 & 0&80\tabularnewline
\hline 
10000 & Custom & \textit{ll} & 7&29 & 1&20 & 0&62 & 0&80\tabularnewline
\hline 
10000 & Custom & \textit{lllat} & 181& & 1&20 & 0&62 & 0&80\tabularnewline
\hline 
\end{tabular}
\par\end{centering}
\caption{Learning results and times\label{tab:Learning-results-and}}
\end{table}

With latency $\tau=2$, we find that we can still recover the parameters
of the kernel quite well (still 100 paths):

\begin{table}[H]
\begin{centering}
\begin{tabular}{|r|r@{\extracolsep{0pt}.}l|r@{\extracolsep{0pt}.}l|r@{\extracolsep{0pt}.}l|r@{\extracolsep{0pt}.}l|}
\hline 
End Time & \multicolumn{2}{c|}{runtime} & \multicolumn{2}{c|}{$\lambda_{0}$} & \multicolumn{2}{c|}{$\alpha$} & \multicolumn{2}{c|}{$\beta$}\tabularnewline
\hline 
\hline 
100 & 1&76 & 1&41 & 0&59 & 0&81\tabularnewline
\hline 
1000 & 15&6 & 1&26 & 0&62 & 0&80\tabularnewline
\hline 
10000 & 223& & 1&21 & 0&61 & 0&79\tabularnewline
\hline 
\end{tabular}
\par\end{centering}
\caption{Learning results and times for latency $\tau=2$}
\end{table}

\section{Multidimensional Hawkes}

\subsection{No latency}

For $P=1$ and $\lambda_{0}\left(t\right)=\lambda_{0}$ :

\begin{equation}
\lambda^{m}\left(t^{m}\right)=\lambda_{0}^{m}+\sum_{n=1}^{M}\sum_{t_{i}^{n}<t^{m}}\left[\alpha^{m,n}\cdot\exp\left(-\beta^{m,n}\cdot\left(t^{m}-t_{i}^{n}\right)\right)\right]
\end{equation}

With the recursion now:

\begin{equation}
R^{m,n}\left(i\right)=R^{m,n}\left(i-1\right)\cdot\exp\left(-\beta^{m,n}\cdot\left(t_{i}^{m}-t_{i-1}^{m}\right)\right)+\sum_{t_{i-1}^{m}\leq t_{k}^{n}<t_{i}^{m}}\exp\left(-\beta^{m,n}\cdot\left(t_{i}^{m}-t_{k}^{n}\right)\right)
\end{equation}

And log-likelihood for each node m:

\begin{align}
LL^{m} & =t_{N}^{m}\cdot\left(1-\lambda_{0}^{m}\right)-\sum_{n=1}^{M}\sum_{t_{i}^{n}<t_{N}^{m}}\left[\frac{\alpha^{m,n}}{\beta^{m,n}}\left(1-\exp\left(-\beta^{m,n}\cdot\left(t_{N}^{m}-t_{i}^{n}\right)\right)\right)\right]\nonumber \\
 & +\sum_{t_{i}^{n}<t_{N}^{m}}\ln\left[\lambda_{0}^{m}+\sum_{n=1}^{M}\alpha^{m,n}\cdot R^{m,n}\left(i\right)\right]
\end{align}

\subsection{Latency}

For $P=1$ and $\lambda_{0}\left(t\right)=\lambda_{0}$ :

\begin{equation}
\lambda^{m}\left(t^{m}\right)=\lambda_{0}^{m}+\sum_{n=1}^{M}\sum_{t_{i}^{n}<t^{m}-\tau}\left[\alpha^{m,n}\cdot\exp\left(-\beta^{m,n}\cdot\left(t^{m}-\tau-t_{i}^{n}\right)\right)\right]
\end{equation}

With the recursion now:

\begin{equation}
R^{m,n}\left(i\right)=R^{m,n}\left(i-1\right)\cdot\exp\left(-\beta^{m,n}\cdot\left(t_{i}^{m}-t_{i-1}^{m}\right)\right)+\sum_{t_{i-1}^{m}-\tau\leq t_{k}^{n}<t_{i}^{m}-\tau}\exp\left(-\beta^{m,n}\cdot\left(t_{i}^{m}-\tau-t_{k}^{n}\right)\right)
\end{equation}

And log-likelihood for each node m:

\begin{align}
LL^{m} & =t_{N}^{m}\cdot\left(1-\lambda_{0}^{m}\right)-\sum_{n=1}^{M}\sum_{t_{i}^{n}<t_{N}^{m}-\tau}\left[\frac{\alpha^{m,n}}{\beta^{m,n}}\left(1-\exp\left(-\beta^{m,n}\cdot\left(t_{N}^{m}-\tau-t_{i}^{n}\right)\right)\right)\right]\nonumber \\
 & +\sum_{t_{i}^{n}<t_{N}^{m}-\tau}\ln\left[\lambda_{0}^{m}+\sum_{n=1}^{M}\alpha^{m,n}\cdot R^{m,n}\left(i\right)\right]
\end{align}

Now there is a choice on how to efficiently select $t_{k}^{n}$ on
the recursion: either find the corresponding arrays (which could be
empty) for each $\left\{ t_{i-1}^{m},t_{i}^{m}\right\} $ pair with
\textit{numpy.extract} or find the appropriate $t_{i}^{m}$ for each
$t_{k}^{n}$ using \textit{numpy.searchsorted} and building a dictionary;
we chose the latter.

To optimize it further all the arrays are padded so all accesses are
on \textit{numpy.ndarrays} (more details on the code - also see \cite{numpy}).

\begin{algorithm}[H]
\begin{lyxcode}
def~$lllatm\left(\theta,m,nts,t_{N}^{m},\Delta t,\delta t,\tau t\right)$:
\begin{lyxcode}
$\left\{ \lambda_{0_{1}},\lambda_{0_{2}},\ldots,\lambda_{0_{M}},\alpha^{1,1},\alpha^{1,2},\ldots,\alpha^{1,M},\ldots,\alpha^{M,M},\beta^{1,1},\beta^{1,2},\ldots,\beta^{1,M},\ldots,\beta^{M,M}\right\} =\theta$~

parameters~are~defined~as~one~argument~for~the~optimization

$m$~is~the~index~of~the~time~series~in~the~\textit{timestamps}~object

$nts$~is~the~array~of~lengths~of~all~the~time~series~of
\begin{lyxcode}
the~\textit{timestamps~object}
\end{lyxcode}
$t_{N}^{m}$~is~the~last~timestamp~in~all~of~the~time~series~of
\begin{lyxcode}
the~\textit{timestamps~object}
\end{lyxcode}
$\Delta t$~is~the~collection~of~series~$t_{i}^{m}-t_{i-1}^{m}$~for~all~$1\leq m\leq M$~including~$t_{1}^{m}$

$\delta t$~is~the~collection~of~collections~of~series~$t_{N}^{m}-\tau-t_{i-1}^{n}$~
\begin{lyxcode}
for~all~$1\leq m\leq M$~and~$1\leq n\leq M$
\end{lyxcode}
$\tau t$~is~the~collection~of~collections~of~series~of~arrays~$t_{i}^{m}-\tau-t_{k}^{n}$~
\begin{lyxcode}
for~all~$t_{k}^{n}$~such~that~\textrm{$t_{i-1}^{m}-\tau\leq t_{k}^{n}<t_{i}^{m}-\tau$}~for~all~$1\leq m\leq M$~and~$1\leq n\leq M$
\end{lyxcode}
For~each~n:
\begin{lyxcode}
Define~$R^{m,n}$~as~an~array~of~zeros~with~the~same~size~as~$t^{m}$

Keep~its~first~element~$R_{1}^{m,n}=0$

Calculate~$R_{i}^{m,n}=\exp\left(-\beta^{m,n}\cdot\Delta t_{i}^{m}\right)\cdot R_{i-1}^{m,n}+\sum_{\tau t_{i}}\left(\exp\left(-\beta^{m,n}\cdot\tau t_{i}^{m,n}\right)\right)$~
\begin{lyxcode}
recursively~for~$i>1$
\end{lyxcode}
\end{lyxcode}
Return:

\begin{align*}
-\left[t_{N}^{m}\cdot\left(1-\lambda_{0}^{m}\right)-\sum_{\delta t}\left(\left(\frac{\alpha^{m,n}}{\beta^{m,n}}\right)\cdot\left(1-\exp\left(-\beta^{m,n}\cdot\delta t^{m,n}\right)\right)\right)\right]\\
-\left[+\sum_{R_{i}}\left(\log\left(\lambda_{0}^{m}+\alpha^{m,n}\cdot R_{i}^{m,n}\right)\right)\right]
\end{align*}
\end{lyxcode}
-{}-{}-{}-{}-{}-{}-{}-{}-{}-{}-{}-{}-{}-{}-{}-{}-{}-{}-{}-{}-{}-{}-{}-{}-{}-{}-{}-{}-{}-{}-{}-{}-{}-{}-{}-{}-{}-{}-{}-{}-{}-{}-{}-{}-{}-{}-{}-{}-{}-{}-{}-{}-{}-{}-{}-{}-{}-{}-{}-{}-{}-{}-{}-{}-{}-{}-{}-{}-{}-{}-{}-{}-{}-{}-{}-

Minimize~$lllatm\left(\theta,m,nts,t_{N}^{m},\Delta t,\delta t,\tau t\right)$~with:
\begin{lyxcode}
bounds~on~$\lambda_{0},\alpha,\beta$~(all~must~be~positive)

no~constraints~for~just~m
\end{lyxcode}
\end{lyxcode}
\caption{Log-likelihood with latency - multidimensional for one time series
m}
\end{algorithm}

A slower optimization can be done for more than one time series in
parallel, with the advantage of enforcing symmetries on coefficients
by re-defining the input $\theta$ with the same decays for blocks
(e.g. on \cite{Bacry et al 2016} instead of 16 different decays we
can use 4 decays, as described further down the paper); the algorithm
below shows the case where we optimize it for all the time series,
but we can modify it to run on bid-ask pairs of time series.

\begin{algorithm}[H]
\begin{lyxcode}
def~$lllatall\left(\theta,nts,t_{N}^{m},\Delta t,\delta t,\tau t\right)$:
\begin{lyxcode}
$\left\{ \lambda_{0_{1}},\lambda_{0_{2}},\ldots,\lambda_{0_{M}},\alpha^{1,1},\alpha^{1,2},\ldots,\alpha^{1,M},\ldots,\alpha^{M,M},\beta^{1,1},\beta^{1,2},\ldots,\beta^{1,M},\ldots,\beta^{M,M}\right\} =\theta$~

parameters~are~defined~as~one~argument~for~the~optimization

$m$~is~the~index~of~the~time~series~in~the~\textit{timestamps}~object

$nts$~is~the~array~of~lengths~of~all~the~time~series~of~
\begin{lyxcode}
the~\textit{timestamps~object}
\end{lyxcode}
$t_{N}^{m}$~is~the~last~timestamp~in~all~of~the~time~series~of
\begin{lyxcode}
the~\textit{timestamps~object}
\end{lyxcode}
$\Delta t$~is~the~collection~of~series~$t_{i}^{m}-t_{i-1}^{m}$~for~all~$1\leq m\leq M$~including~$t_{1}^{m}$

$\delta t$~is~the~collection~of~collections~of~series~$t_{N}^{m}-\tau-t_{i-1}^{n}$~
\begin{lyxcode}
for~all~$1\leq m\leq M$~and~$1\leq n\leq M$
\end{lyxcode}
$\tau t$~is~the~collection~of~collections~of~series~of~arrays~$t_{i}^{m}-\tau-t_{k}^{n}$~
\begin{lyxcode}
for~all~$t_{k}^{n}$~such~that~\textrm{$t_{i-1}^{m}-\tau\leq t_{k}^{n}<t_{i}^{m}-\tau$}~for~all~$1\leq m\leq M$~and~$1\leq n\leq M$
\end{lyxcode}
For~each~m:
\begin{lyxcode}
For~each~n:
\begin{lyxcode}
Define~$R^{m,n}$~as~an~array~of~zeros~with~the~same~size~as~$t^{m}$

Keep~its~first~element~$R_{1}^{m,n}=0$

Calculate~$R_{i}^{m,n}=\exp\left(-\beta^{m,n}\cdot\Delta t_{i}^{m}\right)\cdot R_{i-1}^{m,n}+\sum_{\tau t_{i}}\left(\exp\left(-\beta^{m,n}\cdot\tau t_{i}^{m,n}\right)\right)$~
\begin{lyxcode}
recursively~for~$i>1$
\end{lyxcode}
\end{lyxcode}
Return:

\begin{align*}
LL^{m} & =\left[t_{N}^{m}\cdot\left(1-\lambda_{0}^{m}\right)-\sum_{\delta t}\left(\left(\frac{\alpha^{m,n}}{\beta^{m,n}}\right)\cdot\left(1-\exp\left(-\beta^{m,n}\cdot\delta t^{m,n}\right)\right)\right)\right]\\
+ & \left[\sum_{R_{i}}\left(\log\left(\lambda_{0}^{m}+\alpha^{m,n}\cdot R_{i}^{m,n}\right)\right)\right]
\end{align*}
\end{lyxcode}
Return~$-\sum_{m}\left(LL^{m}\right)$~
\end{lyxcode}
-{}-{}-{}-{}-{}-{}-{}-{}-{}-{}-{}-{}-{}-{}-{}-{}-{}-{}-{}-{}-{}-{}-{}-{}-{}-{}-{}-{}-{}-{}-{}-{}-{}-{}-{}-{}-{}-{}-{}-{}-{}-{}-{}-{}-{}-{}-{}-{}-{}-{}-{}-{}-{}-{}-{}-{}-{}-{}-{}-{}-{}-{}-{}-{}-{}-{}-{}-{}-{}-{}-{}-{}-{}-{}-{}-

Minimize~$lllatall\left(\theta,nts,t_{N}^{m},\Delta t,\delta t,\tau t\right)$~with:
\begin{lyxcode}
bounds~on~$\lambda_{0},\alpha,\beta$~(all~must~be~positive)

inequality~constraints:~stationarity~condition

equality~constraints:~symmetries~on~coefficients
\end{lyxcode}
\end{lyxcode}
\caption{Log-likelihood with latency - multidimensional for all time series}
\end{algorithm}

\subsection{Simulations and learning}

With latency $\tau=2$ and two kernels we get good results (100 paths,
method \textcolor{red}{'SLSQP'}):

\begin{table}[H]
\begin{centering}
\begin{tabular}{|r|r@{\extracolsep{0pt}.}l|r@{\extracolsep{0pt}.}l|r@{\extracolsep{0pt}.}l|r@{\extracolsep{0pt}.}l|r@{\extracolsep{0pt}.}l|r@{\extracolsep{0pt}.}l|r@{\extracolsep{0pt}.}l|r@{\extracolsep{0pt}.}l|r@{\extracolsep{0pt}.}l|r@{\extracolsep{0pt}.}l|r@{\extracolsep{0pt}.}l|}
\hline 
End Time & \multicolumn{2}{c|}{runtime(s)} & \multicolumn{2}{c|}{$\lambda_{0}^{1}$} & \multicolumn{2}{c|}{$\lambda_{0}^{2}$} & \multicolumn{2}{c|}{$\alpha^{1,1}$} & \multicolumn{2}{c|}{$\alpha^{1,2}$} & \multicolumn{2}{c|}{$\alpha^{2,1}$} & \multicolumn{2}{c|}{$\alpha^{2,2}$} & \multicolumn{2}{c|}{$\beta^{1,1}$} & \multicolumn{2}{c|}{$\beta^{1,2}$} & \multicolumn{2}{c|}{$\beta^{2,1}$} & \multicolumn{2}{c|}{$\beta^{2,2}$}\tabularnewline
\hline 
\multicolumn{3}{|r|}{Simulated} & 0&6 & 0&2 & 0&5 & 0&7 & 0&9 & 0&3 & 1&4 & 1&8 & 2&2 & 1&0\tabularnewline
\hline 
\hline 
100 & 7&1 & 0&68 & 0&28 & 0&55 & 0&73 & 1&03 & 0&37 & 1&87 & 1&93 & 2&44 & 2&65\tabularnewline
\hline 
1000 & 67& & 0&63 & 0&21 & 0&52 & 0&75 & 0&98 & 0&31 & 1&44 & 1&78 & 2&18 & 1&05\tabularnewline
\hline 
10000 & 804& & 0&60 & 0&20 & 0&52 & 0&75 & 0&97 & 0&31 & 1&37 & 1&76 & 2&12 & 1&00\tabularnewline
\hline 
\end{tabular}
\par\end{centering}
\caption{Learning results and times for latency $\tau=2$}
\end{table}

\section{Results on real data}

\subsection{Estimation of Slowly Decreasing Hawkes Kernels: Application to High
Frequency Order Book Dynamics}

Here one must consider the latency of $250\mu s$ (as informed in
\cite{Eurex 2017}):

\begin{figure}[H]
\begin{centering}
\includegraphics[scale=0.3]{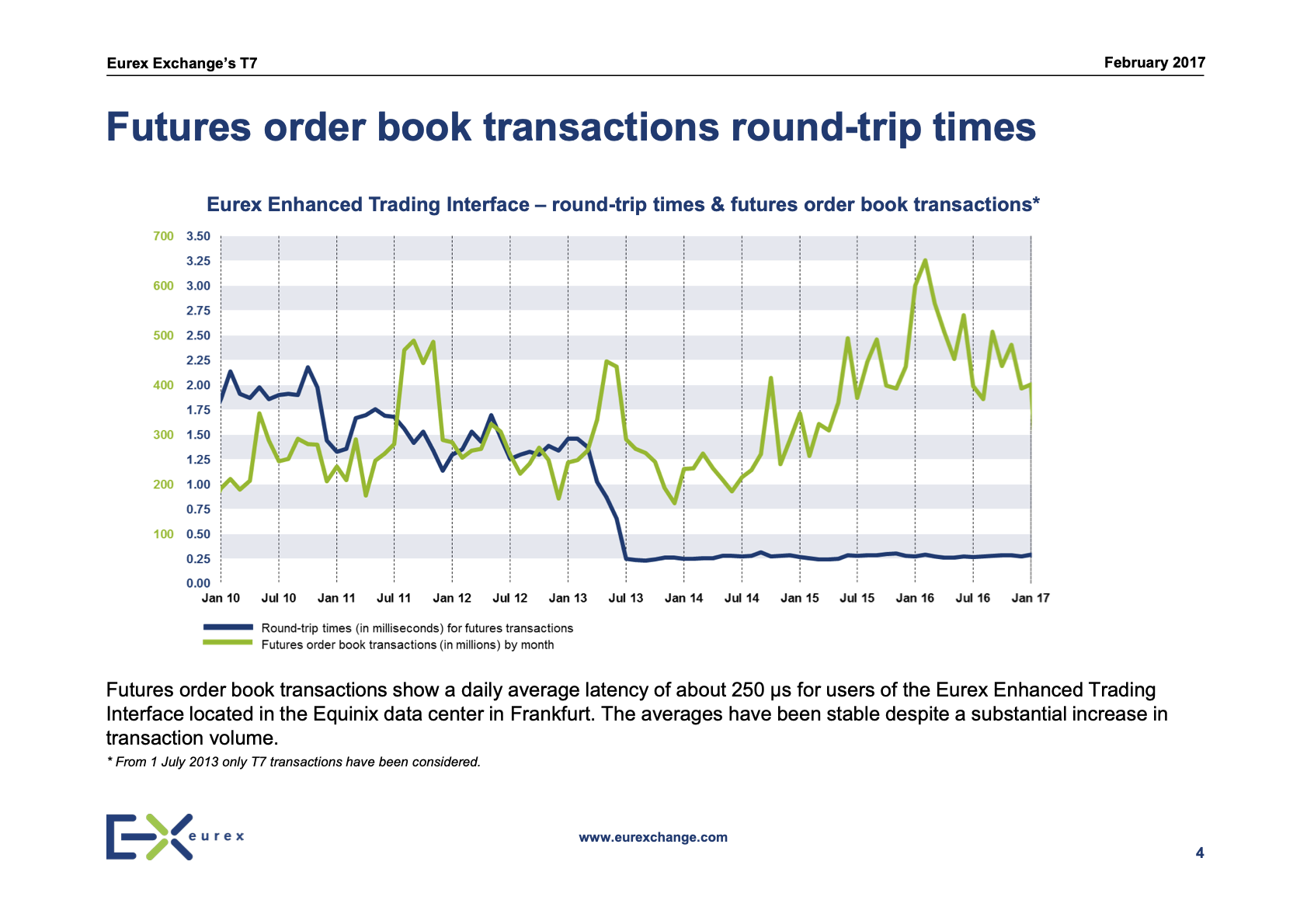}
\par\end{centering}
\caption{Eurex Latency}

\end{figure}

We have available data for 20 days (Apr-2014) on events P and T, from
8h to 22h (14 hours on each day), as described in \url{https://github.com/X-DataInitiative/tick-datasets/tree/master/hawkes/bund}.
Because in some days we have late starts, and because liquidity late
at night is not better, we consider events only from 10h to 18h, which
represent more than 75\% of all events (daily counts available on
GitHub).

To get a better idea of the time scales involved, we calculate the
descriptive statistics of each time series:

\begin{table}[H]
\begin{centering}
\begin{tabular}{|r|r|r|r|r|}
\hline 
 & $P_{u}$ & $P_{d}$ & $T_{a}$ & $T_{b}$\tabularnewline
\hline 
\hline 
$\overline{n}$ (events) & 4569 & 4558 & 11027 & 12811\tabularnewline
\hline 
$\Delta t_{10\%}$ (\textgreek{m}s) & 196 & 196 & 147 & 229\tabularnewline
\hline 
$\Delta t_{15\%}$ (\textgreek{m}s) & 241 & 242 & 317 & 471\tabularnewline
\hline 
$\Delta t_{25\%}$ (\textgreek{m}s) & 342 & 340 & 1890 & 11756\tabularnewline
\hline 
$\Delta t_{50\%}$ (ms) & 14.1 & 13.0 & 159.7 & 183.2\tabularnewline
\hline 
$\Delta t_{75\%}$ (s) & 2.5 & 2.4 & 1.9 & 1.5\tabularnewline
\hline 
$\Delta t_{90\%}$ (s) & 16.9 & 17.1 & 7.5 & 6.3\tabularnewline
\hline 
$\overline{\Delta t}$ (s) & 6.3 & 6.3 & 2.6 & 2.2\tabularnewline
\hline 
\end{tabular}
\par\end{centering}
\caption{Statistics for the time intervals within each series}

\end{table}

And we can see that between 10\% and 15\% of events within each time
series happen less than $250\mu s$ after the previous event, and
therefore should not be considered as being caused by that previous
event. We can also see that the average time interval is considerably
higher than the median time interval, which means that this distribution
is more fat-tailed than an exponential distribution.

We run optimizations for the sum of (Bid+Ask) log-likelihoods where
the kernel intensities are the same within diagonals: $\alpha\left(T_{a}\rightarrow P_{u}\right)=\alpha\left(T_{b}\rightarrow P_{d}\right)$
and $\alpha\left(T_{b}\rightarrow P_{u}\right)=\alpha\left(T_{a}\rightarrow P_{d}\right)$
and the kernel decays are the same within blocks $\beta\left(T_{a}\rightarrow P_{u}\right)=\beta\left(T_{b}\rightarrow P_{d}\right)=\beta\left(T_{b}\rightarrow P_{u}\right)=\beta\left(T_{a}\rightarrow P_{d}\right)$;
so the total LL is optimized individually as $LL^{opt}=LL_{P_{a}+P_{b}}^{opt}+LL_{T_{a}+T_{b}}^{opt}$.
We also run exploratory optimizations with completely independent
intensities and decays, and tested different optimizer algorithms,
in order to reduce the bounds for the parameters. The code (including
initial values and bounds) is at the GitHub page, together with the
spreadsheet with the results of the optimization (using the Differential
Evolution, Powell and L-BFGS-B methods from \cite{scipy}; Powell
was the best-performing method). The parameters for $T\rightarrow T$
were the most unstable (charts in the Appendix).

\begin{table}[H]
\begin{centering}
\begin{tabular}{|c|c|c|c|c|}
\hline 
 & $P_{u}$ & $P_{d}$ & $T_{a}$ & $T_{b}$\tabularnewline
\hline 
\hline 
$\lambda_{0}$ & 4.0\% & 4.0\% & 14.3\% & 16.1\%\tabularnewline
\hline 
R & 25\% & 25\% & 38\% & 37\%\tabularnewline
\hline 
\end{tabular}
\par\end{centering}
\vphantom{}

\vphantom{}

\vphantom{}
\begin{centering}
\begin{tabular}{|c|c|c|c|c|}
\hline 
 & $P_{u}\rightarrow P_{u}$ & $P_{d}\rightarrow P_{u}$ & $T_{a}\rightarrow P_{u}$ & $T_{b}\rightarrow P_{u}$\tabularnewline
\hline 
\hline 
$\alpha$ & 737 & 488 & 401 & 14.5\tabularnewline
\hline 
$\beta$ & \multicolumn{2}{c|}{3113} & \multicolumn{2}{c|}{3125}\tabularnewline
\hline 
$\alpha/\beta$ & 23.7\% & 15.7\% & 12.8\% & 0.46\%\tabularnewline
\hline 
\end{tabular}
\par\end{centering}
\vphantom{}

\vphantom{}

\vphantom{}
\begin{centering}
\begin{tabular}{|c|c|c|c|c|}
\hline 
 & $P_{u}\rightarrow T_{a}$ & $P_{d}\rightarrow T_{a}$ & $T_{a}\rightarrow T_{a}$ & $T_{b}\rightarrow T_{a}$\tabularnewline
\hline 
\hline 
$\alpha$ & 23.6 & 316 & 3.8 (median) & 0.27 (median)\tabularnewline
\hline 
$\beta$ & \multicolumn{2}{c|}{1122} & \multicolumn{2}{c|}{7.8 (median)}\tabularnewline
\hline 
$\alpha/\beta$ & 2.1\% & 28.3\% & 48.0\% & 3.1\%\tabularnewline
\hline 
\end{tabular}
\par\end{centering}
\caption{Learned parameters}
\end{table}

We calculate the exogeneity ratio R between the number of exogenous
events and the total number of events n for each event type i and
day d as:

\begin{equation}
R_{i,d}=\frac{\left(\lambda_{0}\right)_{i,d}}{\frac{\left(n\right)_{i,d}}{8\cdot3600}}
\end{equation}

\begin{figure}[H]
\begin{centering}
\includegraphics[scale=0.5]{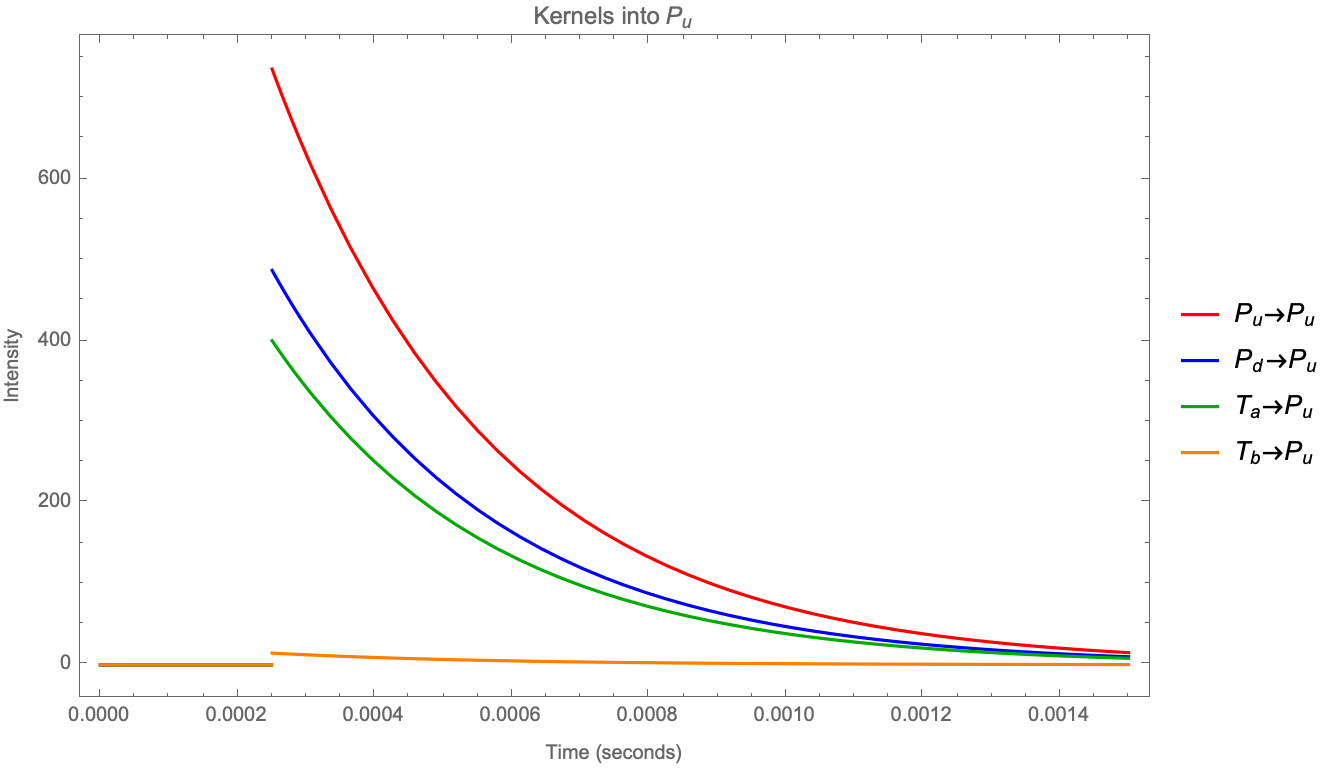}
\par\end{centering}
\begin{centering}
\includegraphics[scale=0.5]{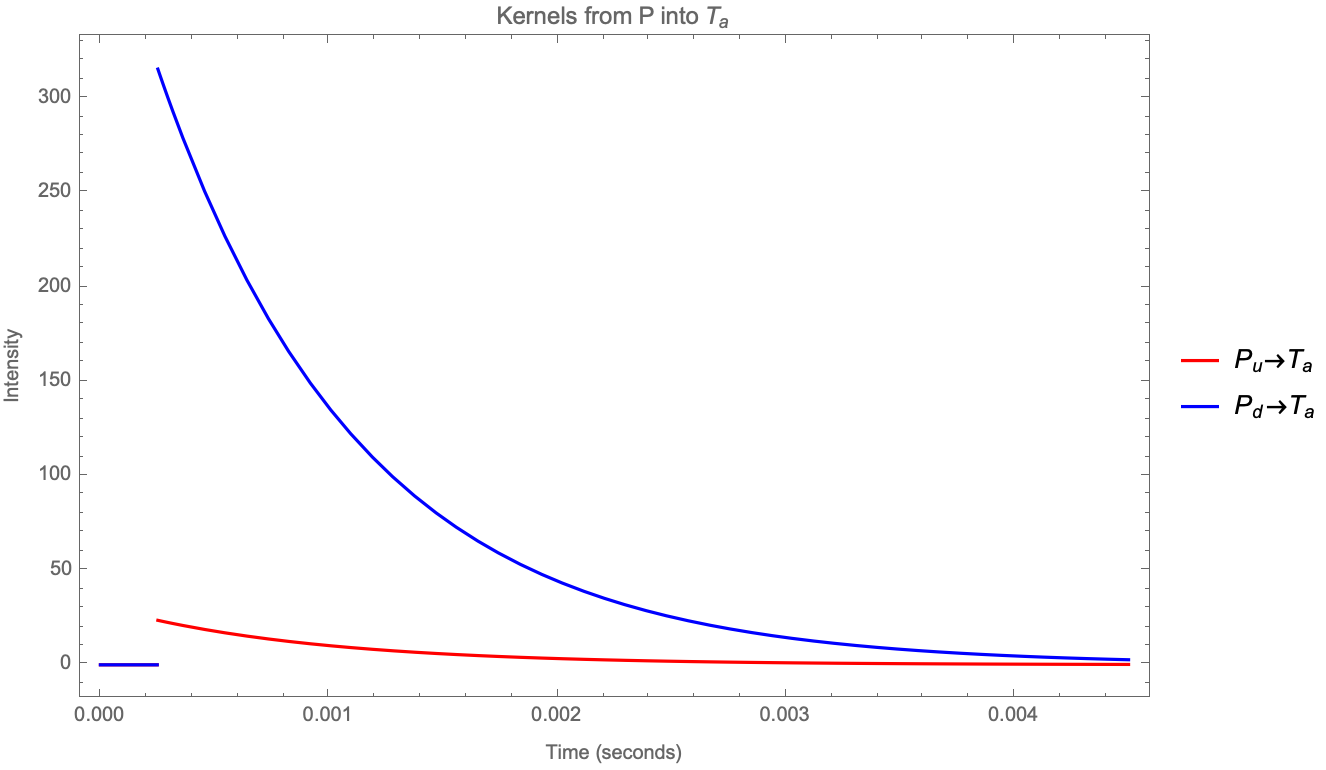}
\par\end{centering}
\begin{centering}
\includegraphics[scale=0.5]{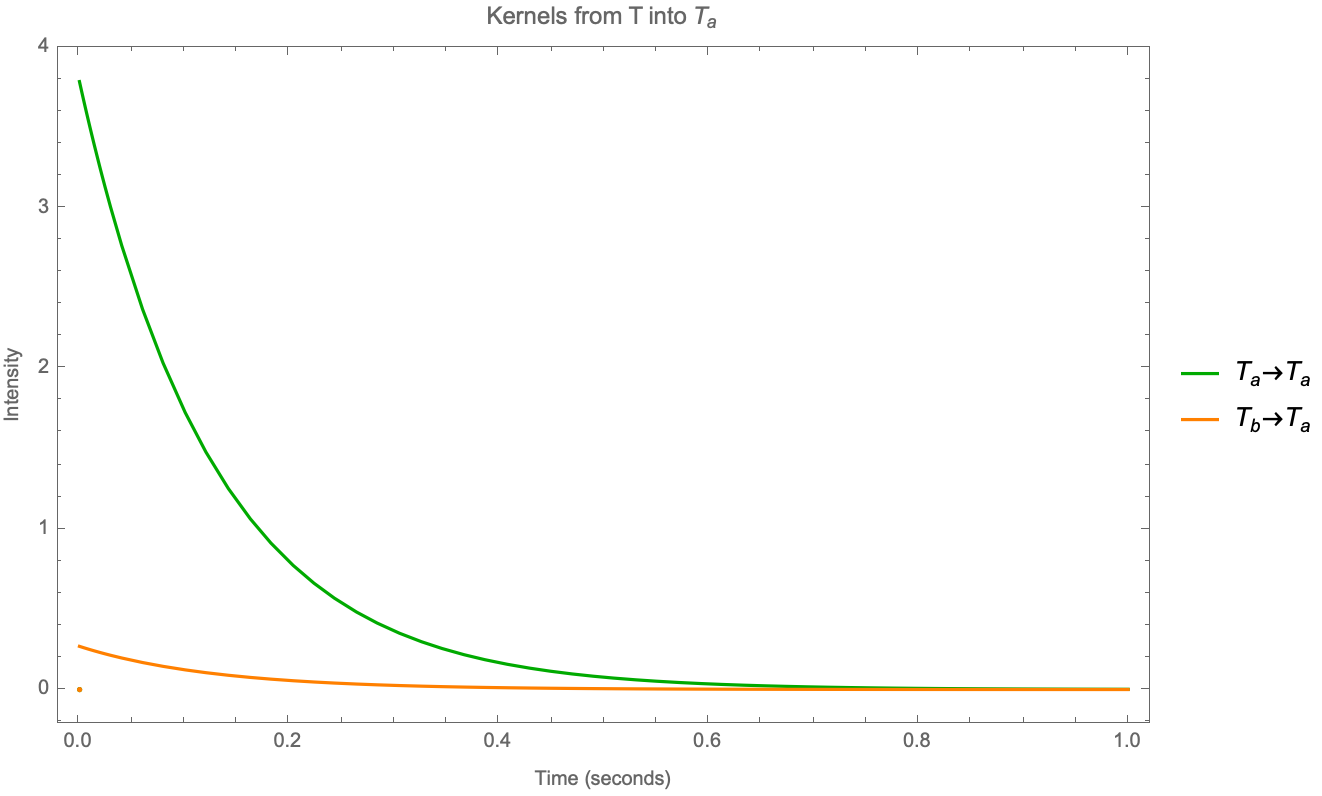}
\par\end{centering}
\caption{Learned Exponential Kernels\label{fig:Learned-Exponential-Kernels}}
\end{figure}

And these ratios are much higher than those found in \cite{Bacry et al 2016},
which can be explained as the reassignment of the events within the
latency. Even then we find that it is reasonable to expect that a
good part of the trades are exogenous (ie caused by reasons other
than short-term dynamics of the order book). Most interestingly, events
that change the mid-price (trades, cancels and replenishments) are
(i) faster and (ii) more endogenous than trades that do not change
the mid-price. Please note how the first chart of Figure \ref{fig:Learned-Exponential-Kernels}
is similar to the first chart of Figure 7 in \cite{Bacry et al 2016}
after approximately $250\mu s$.

It's not by chance that the decay on $T\rightarrow P$ and $P\rightarrow P$
kernels is about 3000; we can plot the first chart again but changing
the x-axis to units of latency:

\begin{figure}[H]
\begin{centering}
\includegraphics[scale=0.6]{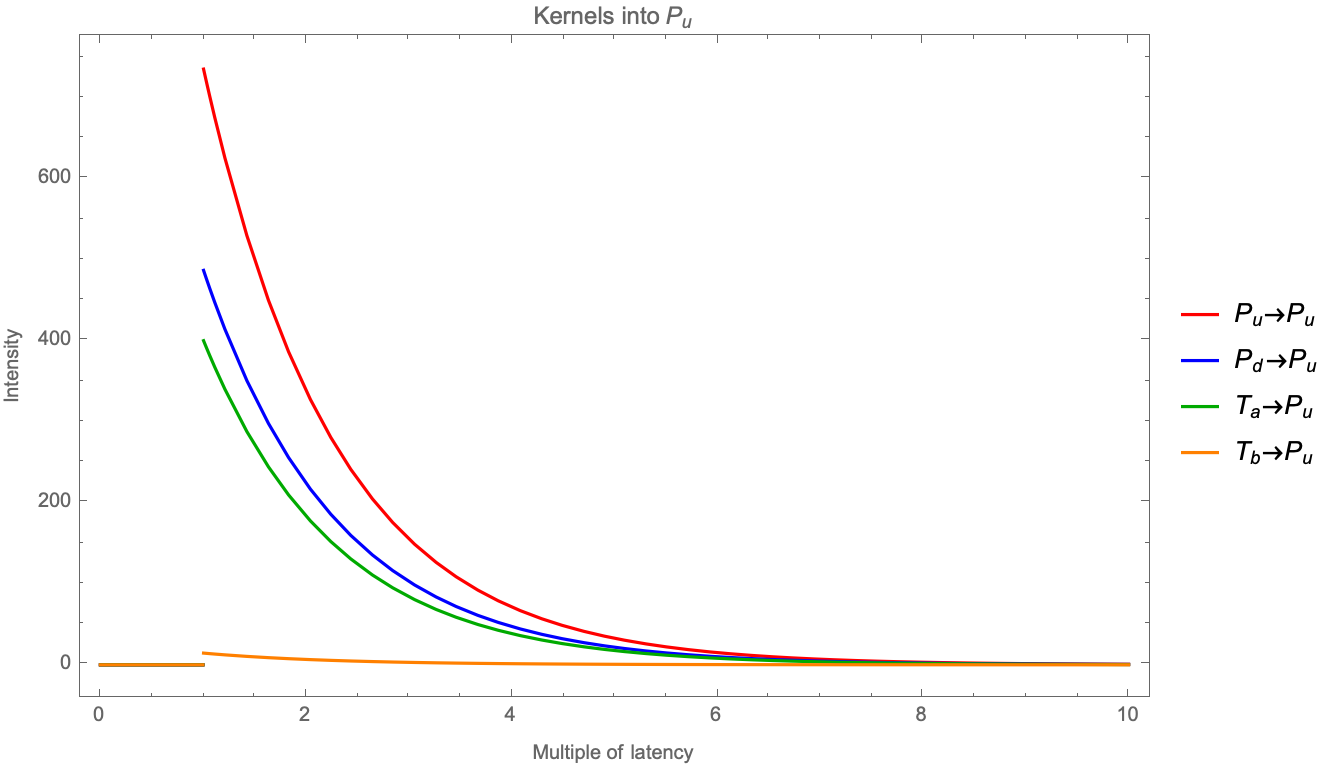}
\par\end{centering}
\caption{$P,T\rightarrow P$ Kernels in units of latency\label{fig:-Kernels-in-units}}

\end{figure}

It's expected that fast participants will act quickly on: (i) sharp
imbalances by either canceling or consuming the remaining queue $\left(T_{a}\rightarrow P_{u}\right)$
(ii) filling in recent gaps $\left(P\rightarrow P\right)$ and that
they will be able to react to the reactions as quickly as before,
so the influence of the first event decays fast. We expect that Figure
\ref{fig:-Kernels-in-units} will be found in different markets and
different years (e.g. BUND futures in 2012, which had a latency of
$1250\mu s$ instead of $250\mu s$).

The main differences found with \cite{Bacry et al 2016} are on the
anti-diagonals. For $P\rightarrow P$ we find that diagonals are stronger
than anti-diagonals (even on a large tick asset such as the BUND future);
to investigate this further a better breakdown of P into Cancels and
Trades would help. For $P\rightarrow T$ our method is not able to
learn negative intensities because of the log-likelihood, but none
of the values reached the lower bound; average intensity was about
2\%; although a change in mid-price will make limit orders with the
same sign miss the fill and become diagonals in the sub-matrix $P\rightarrow P$
market orders would not be affected; but more importantly the speed
and intensity of the $P\rightarrow P$ kernels ensure that gaps close
quickly; we think that the introduction of the latency is enough to
simplify the model such that negative intensities would not be necessary.

We have not tested a Sum of Exponentials model, but it might be interesting
to fix the decays for the first exponential kernel at ``known''
values and see what we can find.

\section{Conclusions}

We show (with formulas and code) how to simulate and estimate exponential
kernels of Hawkes processes with a latency, and interpreted the results
of applying this method to a (limited) set of BUND futures data. We
show how some features of this model (the $P\rightarrow P$ kernels
and parameter symmetries) might be quite universal in markets. We
can also extend this approach to two related securities that trade
in different markets by using different latencies on the cross-market
kernels without changing the model.

We would like to thank Mathieu Rosenbaum (CMAP) for the guidance and
Sebastian Neusuess (Eurex) for the latency data.

\pagebreak{}

\appendix

\part*{Appendix}

\section{Appendix}

The charts and tables for the daily estimates are below. The only
real worry is the instability of the parameters of the $T\rightarrow T$
kernels, although the intensities $\frac{\alpha}{\beta}$ are stable.
The upper bound for $\beta\left(T\rightarrow T\right)$ was hit on
day 16.

\begin{figure}[H]
\begin{centering}
\includegraphics[scale=0.3]{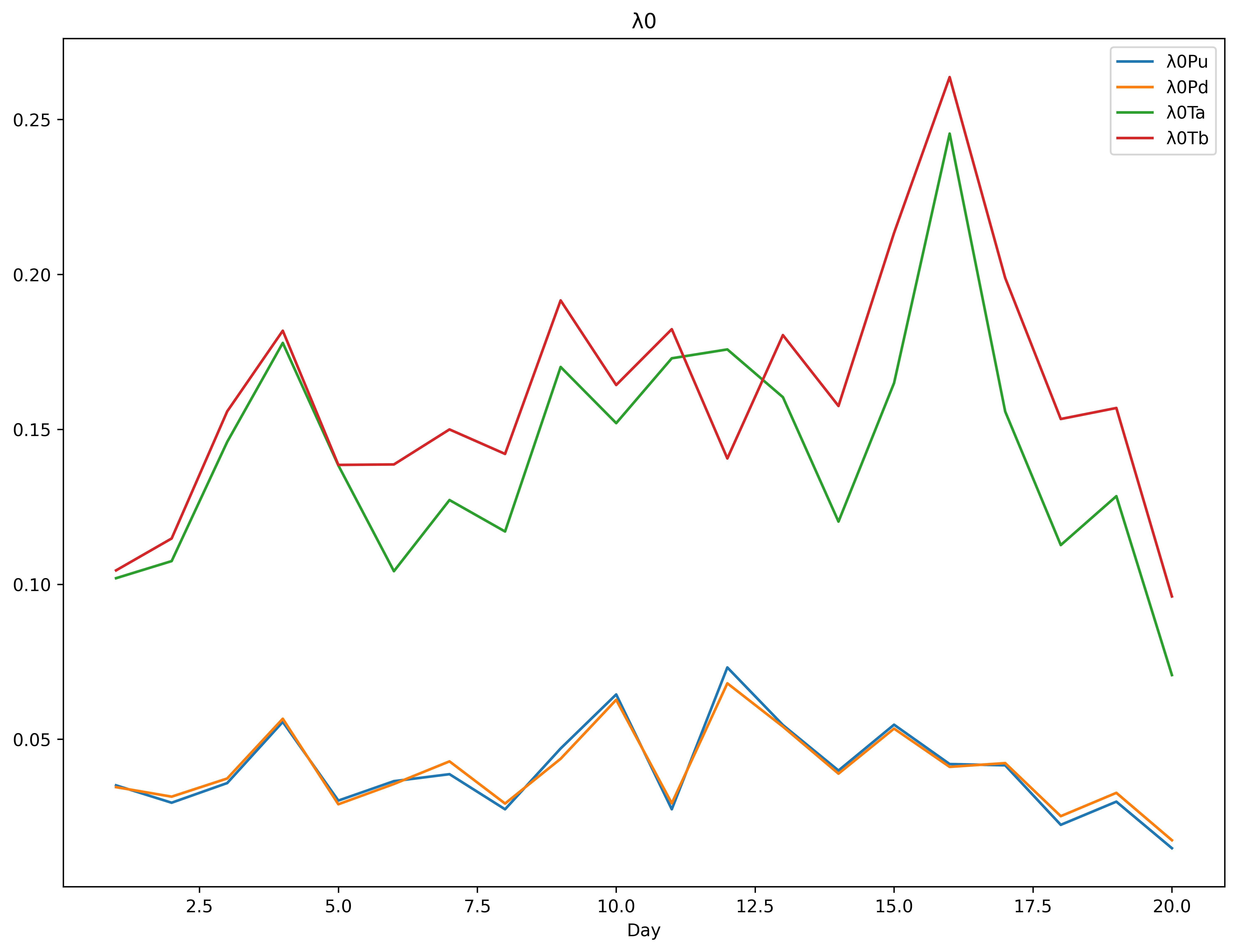}\includegraphics[scale=0.3]{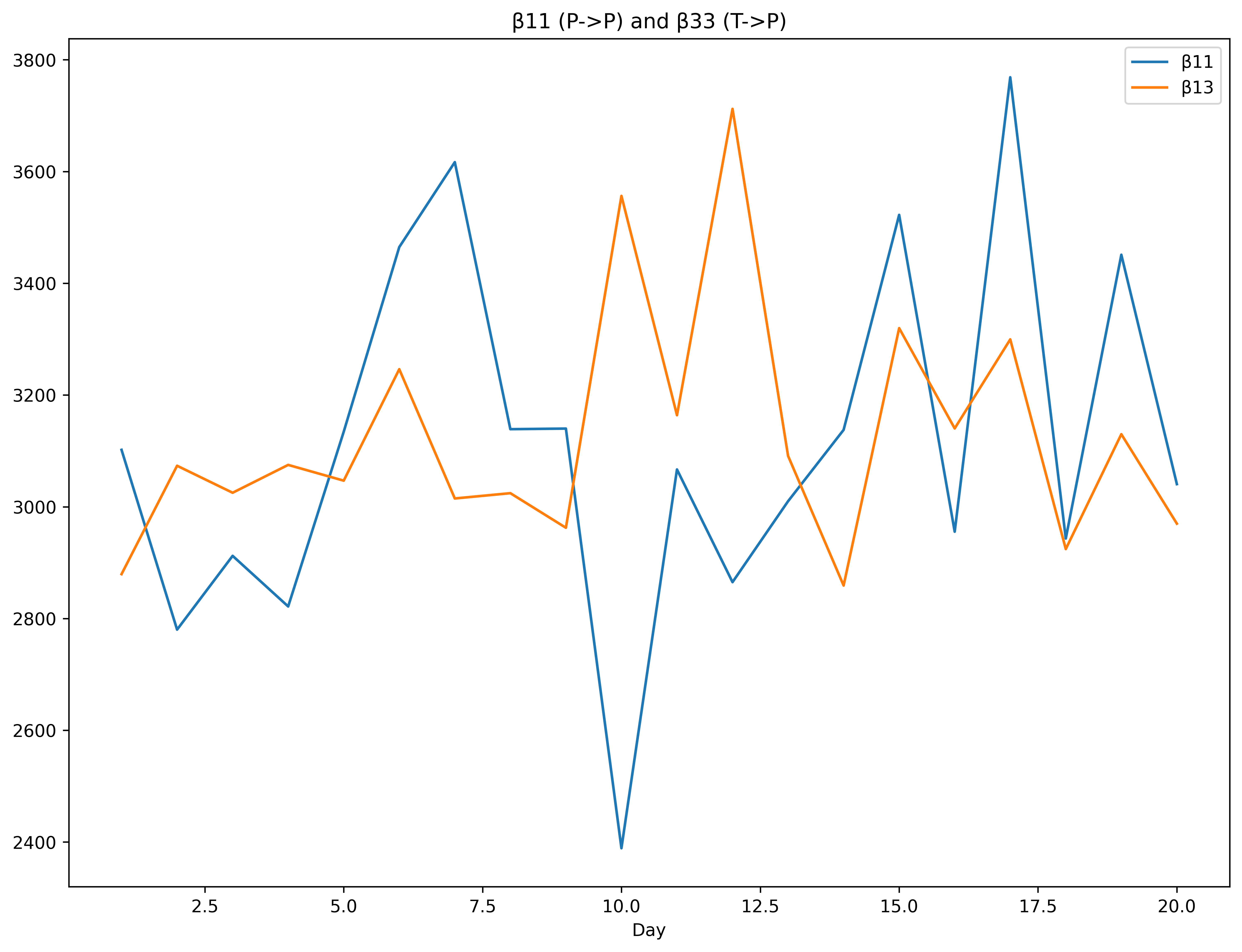}
\par\end{centering}
\begin{centering}
\includegraphics[scale=0.3]{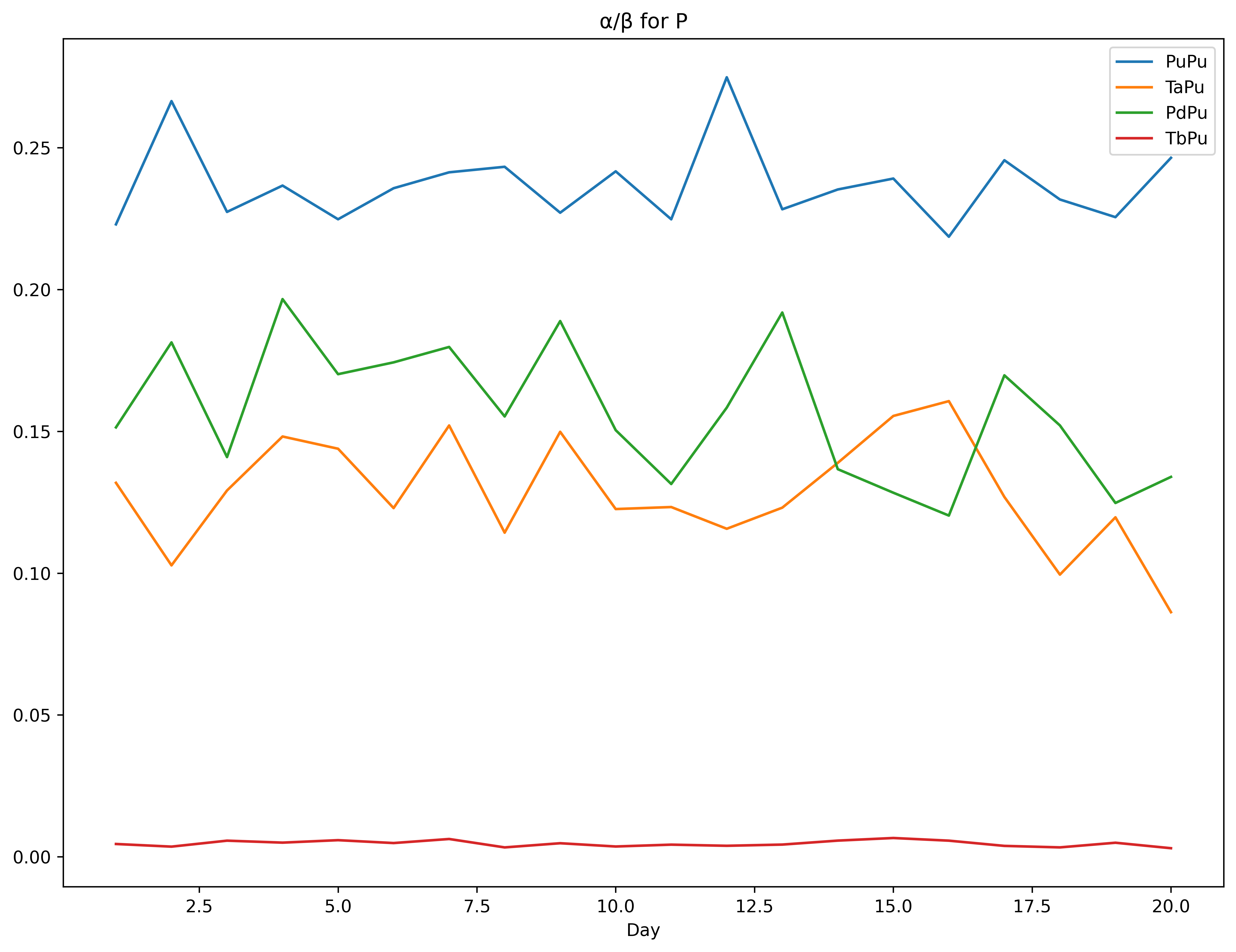}\includegraphics[scale=0.3]{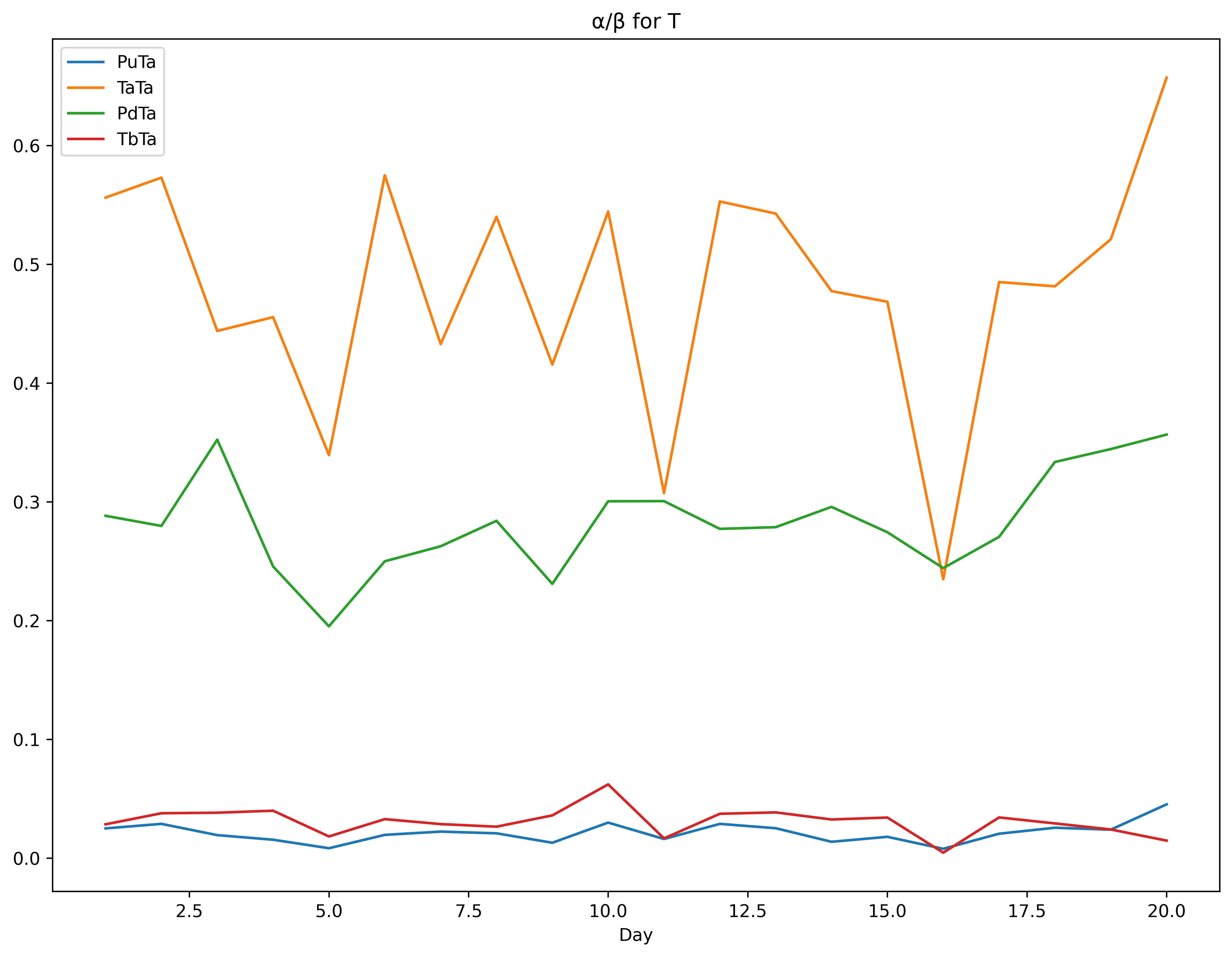}
\par\end{centering}
\caption{Daily estimates of parameters}
\end{figure}

\begin{table}[H]
\begin{centering}
\begin{tabular}{|r|r|r|r|}
\hline 
Day & $\alpha\left(T_{a}\rightarrow T_{a}\right)$ & $\alpha\left(T_{b}\rightarrow T_{a}\right)$ & $\beta\left(T\rightarrow T\right)$\tabularnewline
\hline 
\hline 
1 & 2.73 & 0.14 & 4.91\tabularnewline
\hline 
2 & 2.80 & 0.18 & 4.88\tabularnewline
\hline 
3 & 3.81 & 0.33 & 8.57\tabularnewline
\hline 
4 & 7.12 & 0.62 & 15.64\tabularnewline
\hline 
5 & 36.12 & 1.94 & 106.44\tabularnewline
\hline 
6 & 2.96 & 0.17 & 5.15\tabularnewline
\hline 
7 & 4.68 & 0.31 & 10.81\tabularnewline
\hline 
8 & 4.19 & 0.21 & 7.75\tabularnewline
\hline 
9 & 7.78 & 0.67 & 18.73\tabularnewline
\hline 
10 & 2.55 & 0.29 & 4.68\tabularnewline
\hline 
11 & 22.68 & 1.22 & 73.79\tabularnewline
\hline 
12 & 2.28 & 0.15 & 4.13\tabularnewline
\hline 
13 & 3.33 & 0.24 & 6.14\tabularnewline
\hline 
14 & 4.04 & 0.27 & 8.46\tabularnewline
\hline 
15 & 4.75 & 0.35 & 10.15\tabularnewline
\hline 
16 & 117.41 & 2.28 & \textbf{500.00}\tabularnewline
\hline 
17 & 3.85 & 0.27 & 7.93\tabularnewline
\hline 
18 & 3.63 & 0.22 & 7.54\tabularnewline
\hline 
19 & 3.24 & 0.15 & 6.22\tabularnewline
\hline 
20 & 1.76 & 0.04 & 2.68\tabularnewline
\hline 
\end{tabular}
\par\end{centering}
\caption{Daily estimates of parameters for $T\rightarrow T$ kernels}

\end{table}

\end{document}